\documentclass[conference]{IEEEtran}
\IEEEoverridecommandlockouts
\usepackage{cite}
\usepackage{amsmath,amssymb,amsfonts}
\usepackage{algorithmic}
\usepackage{graphicx}
\usepackage{textcomp}
\usepackage{subfigure}
\usepackage{xcolor}
\usepackage[hidelinks]{hyperref}
\usepackage{soul} %
\usepackage{amsmath} %
\usepackage{tablefootnote} %
\def\BibTeX{{\rm B\kern-.05em{\sc i\kern-.025em b}\kern-.08em
    T\kern-.1667em\lower.7ex\hbox{E}\kern-.125emX}}

\begin{document}

\title{
Backchannel Detection and Agreement Estimation from Video with Transformer Networks
}

\author{Ahmed Amer$^1$, Chirag Bhuvaneshwara$^1$, Gowtham K. Addluri$^2$, Mohammed M. Shaik,$^2$, Vedant Bonde$^2$, \\ Philipp M\"uller$^1$\\
$^1$Cognitive Assistants, German Research Center for Artificial Intelligence, Germany\\ 
$^2$Department of Computer Science, Saarland University, Germany\\
\texttt{\small\{
\href{mailto:ahmed.amer@dfki.de}{ahmed.amer},
\href{mailto:chirag.bhuvaneshwara@dfki.de}{chirag.bhuvaneshwara},
\href{mailto:philipp.mueller@dfki.de}{philipp.mueller}
\}@dfki.de} \\
\texttt{\small\{
\href{mailto:goad00002@stud.uni-saarland.de}{goad00002},
\href{mailto:mosh00003@stud.uni-saarland.de}{mosh00003},
\href{mailto:vebo00001@stud.uni-saarland.de}{vebo00001}
\}@stud.uni-saarland.de}
}

\maketitle

\begin{abstract}
Listeners use short interjections, so-called backchannels, to signify attention or express agreement. 
The automatic analysis of this behavior is of key importance for human conversation analysis and interactive conversational agents.
Current state-of-the-art approaches for backchannel analysis from visual behavior make use of two types of features: features based on body pose and features based on facial behavior.
At the same time, transformer neural networks have been established as an effective means to fuse input from different data sources, but they have not yet been applied to backchannel analysis. 
In this work, we conduct a comprehensive evaluation of multi-modal transformer architectures for automatic backchannel analysis based on pose and facial information.
We address both the detection of backchannels as well as the task of estimating the agreement expressed in a backchannel.
In evaluations on the MultiMediate'22 backchannel detection challenge, we reach 66.4\% accuracy with a one-layer transformer architecture, outperforming the previous state of the art.
With a two-layer transformer architecture, we furthermore set a new state of the art (0.0604 MSE) on the task 
of estimating the amount of agreement expressed in a backchannel.

\end{abstract}


\section{Introduction}

In a conversation, listeners produce backchannels that can consist of a variety of behavioral cues, including vocalizations (e.g. ``yeah'', ``hm'') as well as gestures and facial behavior (e.g. hand movements or nods).
Backchannels are crucial for a smooth conversation, as they are an effective means to communicate attention~\cite{park_telling_2017, kendon_functions_1967}, engagement~\cite{goswami_towards_2020}, and agreement or disagreement while another person is speaking~\cite{kendon_functions_1967, cutrone_backchannel_nodate}. 
They even allow the speaker to adjust their storytelling~\cite{tolins_addressee_2014} and are used by humans
for dialogue comprehension~\cite{tolins2016overhearers} and turn-taking negotiation~\cite{schegloff_discourse_1982}.
Therefore, any artificial system that is supposed to fully understand and effectively support human conversation requires the ability to accurately detect and interpret human backchanneling behavior.

Despite the importance of automatic detection and interpretation of backchanneling behavior, research in this area is still limited~\cite{muller2022multimediate,sharma2022graph}.
These approaches showed that visual behavior observable on the face and in the body posture is highly informative of backchanneling behavior, opening up the possibility for backchannel analysis based on nonverbal visual cues exclusively.
As argued in previous works \cite{HeadPoseEmergentLeaderPred,UpperBodyPoseVAD}, such behavior analysis, based on visual cues only, has the advantage to be robust to low-quality or even unavailable audio recordings. 
However, how to best merge the visual feature representations, obtained from facial behavior with those from body posture, in a single prediction model is not yet well explored.
Previous work on backchannel analysis exclusively used early concatenation to merge feature representations~\cite{muller2022multimediate,sharma2022graph}, as did previous work on engagement and emotion prediction based on postural and facial behavior~\cite{FacePoseFusionEmotionRecog,FacePoseFusionEngagementIntensity}.

In recent years, transformer neural networks have gained popularity due to their breakthrough performance in natural language processing~\cite{AttentionIsAllYouNeed}, audio signal processing~\cite{ASPtransformer} and computer vision~\cite{CVTransformer} tasks.
Transformers proved to be especially successful when fusing several input modalities or feature representations, and several possible architectures for such fusion were proposed~\cite{MultimodalTransformersSurvey}. 
Despite having these suggestions in the literature, research using transformers to fuse different feature representations commonly does not systematically evaluate these different fusion approaches against each other~\cite{multimodalEmotionRecogTransformer,PersonalityAndBodyLanguageRecogTransformer, ValenceArousalEstimationMMTransformer,DyadFormer}.

In our work, we provide a comprehensive evaluation of different transformer architectures for multi-modal fusion for the task of backchannel detection and agreement estimation from backchannels in the recent MultiMediate challenge~\cite{muller2022multimediate}.
This challenge is based on the MPIIGroupInteraction dataset~\cite{muller_detecting_2018} which consists of group discussions of three to four people.
Evaluating all applicable architectures discussed in a recent survey on multi-modal transformers~\cite{MultimodalTransformersSurvey}, we identify that a single transformer layer working jointly on pose- and face-based features outperforms the previous state-of-the-art on backchannel detection.
Similarly, for the task of agreement estimation from backchannels, we set a new state-of-the-art with a vertical stack of two transformer layers. %

\section{Related Work}

Our work is related to computational approaches concerned with human backchannels as well as to transformer-based multi-modal fusion approaches.

\subsection{Automatic Analysis of Backchannels}
Computational approaches addressing the phenomenon of backchanneling in human conversations can be grouped into two categories. 
First, approaches that try to anticipate backchannel insertions~\cite{hara_prediction_2018, adiba_towards_2021, dharo_delay_2021,terrell_regression-based_2012}.
These approaches analyse the conversation in order to find opportune moments to insert backchannels, and are commonly based on word embeddings~\cite{adiba_towards_2021, dharo_delay_2021}, prosodic features~\cite{hara_prediction_2018, adiba_towards_2021}, or part-of-speech-tags and discourse features~\cite{blache_integrated_2020, kawahara_prediction_2016, boudin_multimodal_2021}.
As such, they are not based on the analysis of observed backchannels, but on the analysis of the context that precedes the backchannel.
The second group of approaches addresses the detection of backchannels~\cite{muller2022multimediate,sharma2022graph,saha2020towards}. 
That is, given the observation of a behaviour sequence, these approaches decide whether a backchannel is present in this sequence or not.
Compared to backchannel anticipation, this problem is less well studied.
It can be a part of dialogue act classification~\cite{saha2020towards,alexandersson1997dialogue} based on text and potentially multi-modal input.
Recently, the MultiMediate challenge introduced the task of backchannel detection in group interactions based on audio- and video input~\cite{muller2022multimediate}.
The current state-of-the art approach by Sharma et al.~\cite{sharma2022graph} makes use of graph neural networks to model the relation of individual features inside the discussion group.
While they employ deep audio-visual features~\cite{chung2019perfect}, these yielded only a minor improvement over video-only explicit features computed from body pose~\cite{openpose} and face~\cite{baltrusaitis2018openface} (see ~\autoref{fig:openface-openpose}).
To characterize backchanneling behavior beyond the mere detection of backchannel occurances, the MultiMediate challenge introduced the task of estimating the amount of agreement expressed in a given backchannel~\cite{muller2022multimediate}.
The current state-of-the-art for this task still consists of features based on head pose fed to an SVM classifier~\cite{muller2022multimediate}, and could not be surpassed by graph-based group modelling and deep features~\cite{sharma2022graph}.

To summarize, head- and body pose based features performed well in previous work on backchannel detection and could serve as the basis of a visual-only approach to these tasks which is independent of the availability and quality of audio recordings.
How to best integrate these features into a prediction model however is still underexplored.

\subsection{Multi-modal Fusion with Transformers}
In recent years, transformer neural networks have become increasingly popular as a result of their impressive performance on a variety of domains, including natural language processing~\cite{AttentionIsAllYouNeed}, speech recognition~\cite{ASPtransformer}, and image classification~\cite{CVTransformer}.
The key ingredient of the transformer is the built-in attention mechanism that enables the network to establish relationships between each unit of the input with every other unit.
This inherent feature of transformers makes them especially suited for multi-modal fusion tasks where correspondences between different modalities of feature representations need to be established~\cite{MultimodalTransformersSurvey}.
Multi-modal transformers have been successfully applied to human behavior analysis tasks, including emotion recognition \cite{multimodalEmotionRecogTransformer}, 
personality prediction and body language recognition \cite{DyadFormer,PersonalityAndBodyLanguageRecogTransformer}, as well as valence and arousal estimation \cite{ValenceArousalEstimationMMTransformer}.
While a recent survey on multi-modal transformers by Xu et al.~\cite{MultimodalTransformersSurvey} provided a summary of the most frequently used fusion strategies in transformer networks, the multi-modal transformer architectures in the literature commonly do not provide a principled comparison of these different fusion strategies.
Furthermore, while multi-modal transformers were employed to fuse for example the audio with the video modality, to the best of our knowledge, they were not yet applied to the problem of fusing pose feature representations with facial feature representations~\cite{FacePoseFusionEmotionRecog,FacePoseFusionEngagementIntensity}.

In our work, we conduct the first comprehensive evaluation of different transformer-based fusion mechanisms to process pose- and facial information for the tasks of backchannel detection and agreement estimation from backchannels.

\section{Method}

We first introduce the standard transformer layer from ~\cite{AttentionIsAllYouNeed}.
Subsequently, we explain all transformer architectures for multi-modal fusion that we evaluated for backchannel analysis.
Finally, we detail our feature extraction approach.

\subsection{Transformer Layers}

The transformer layer is an architecture for processing sequential data and was first applied in NLP~\cite{AttentionIsAllYouNeed}.
The key component of the transformer layer is the self-attention mechanism, which allows the model to selectively focus on certain parts of the input sequence while processing it:

\begin{equation}
Attention(Q, K, V) = softmax(\frac{QK^T}{\sqrt{d_k}})V \label{eq:self-attention}
\end{equation}

$\textit{Q}$, $\textit{K}$, and $\textit{V}$ are the query, key and value matrices respectively, which are learned during training. $\textit{d}_{\textit{k}}$ is the dimension of the key matrix. These attention weights are used to compute the weighted sum of the value matrix, which is used as the output of the attention mechanism.

While \autoref{eq:self-attention} describes a single attention head,
transformer layers use multi-headed attention.
This means that the single attention head is repeated $\textit{h}$ times and its output is concatenated.
The result is passed through a linear output transformation using ${\textit{W}}^{\textit{O}}$:

\begin{equation}
MultiHead(Q, K, V) = Concat(head_1, ..., head_h)W^O \label{eq:multi-head-attention}
\end{equation}

The transformer layer also includes positional encoding, which is added to the input embeddings to give the model information about the relative position of the words in the sentence. This allows the model to handle input sequences of varying lengths and to process the order of the elements in the sequence properly. Subsequently, the transformer layer uses multi-layer perceptrons to further process the output of the multi-head self-attention mechanism. 

\begin{figure}
    \centering
    \subfigure{\includegraphics[width=.19\linewidth]{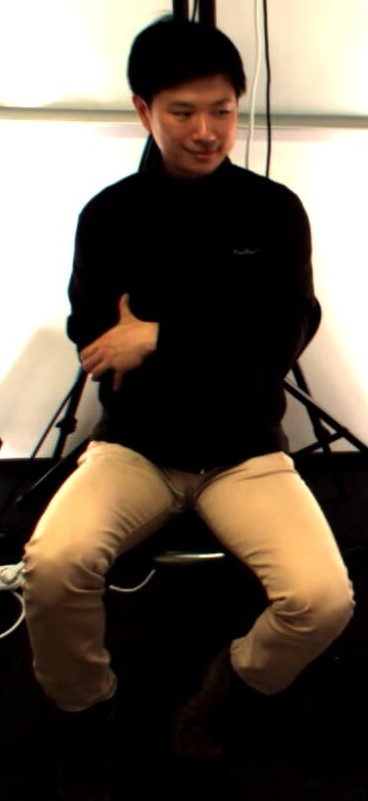}}
    \subfigure{\includegraphics[width=.37\linewidth]{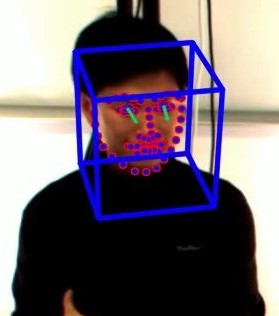}}
    \subfigure{\includegraphics[width=.18\linewidth]{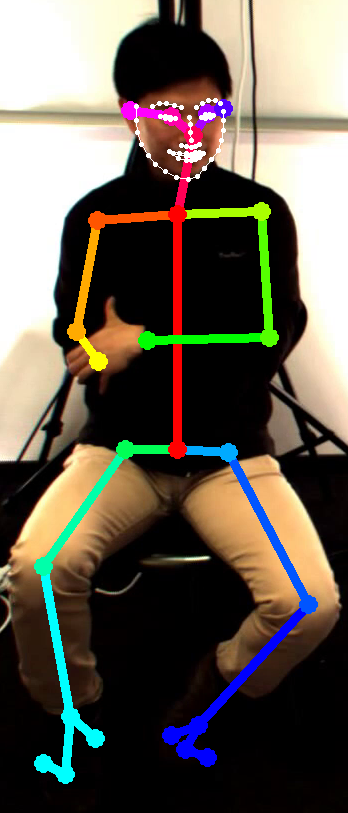}}
    \caption{\textbf{Left:} original frame from the MPIIGroupInteraction dataset \cite{muller_detecting_2018}. \textbf{Middle:} visualization of OpenFace 2.0 \cite{baltrusaitis2018openface} output. \textbf{Right:} visualization of OpenPose \cite{openpose} output. The figures were magnified relative to the original to show the facial features of the participants.}
    \label{fig:openface-openpose}
\end{figure}

\begin{figure*}
  \centering
  \includegraphics[width=\textwidth]{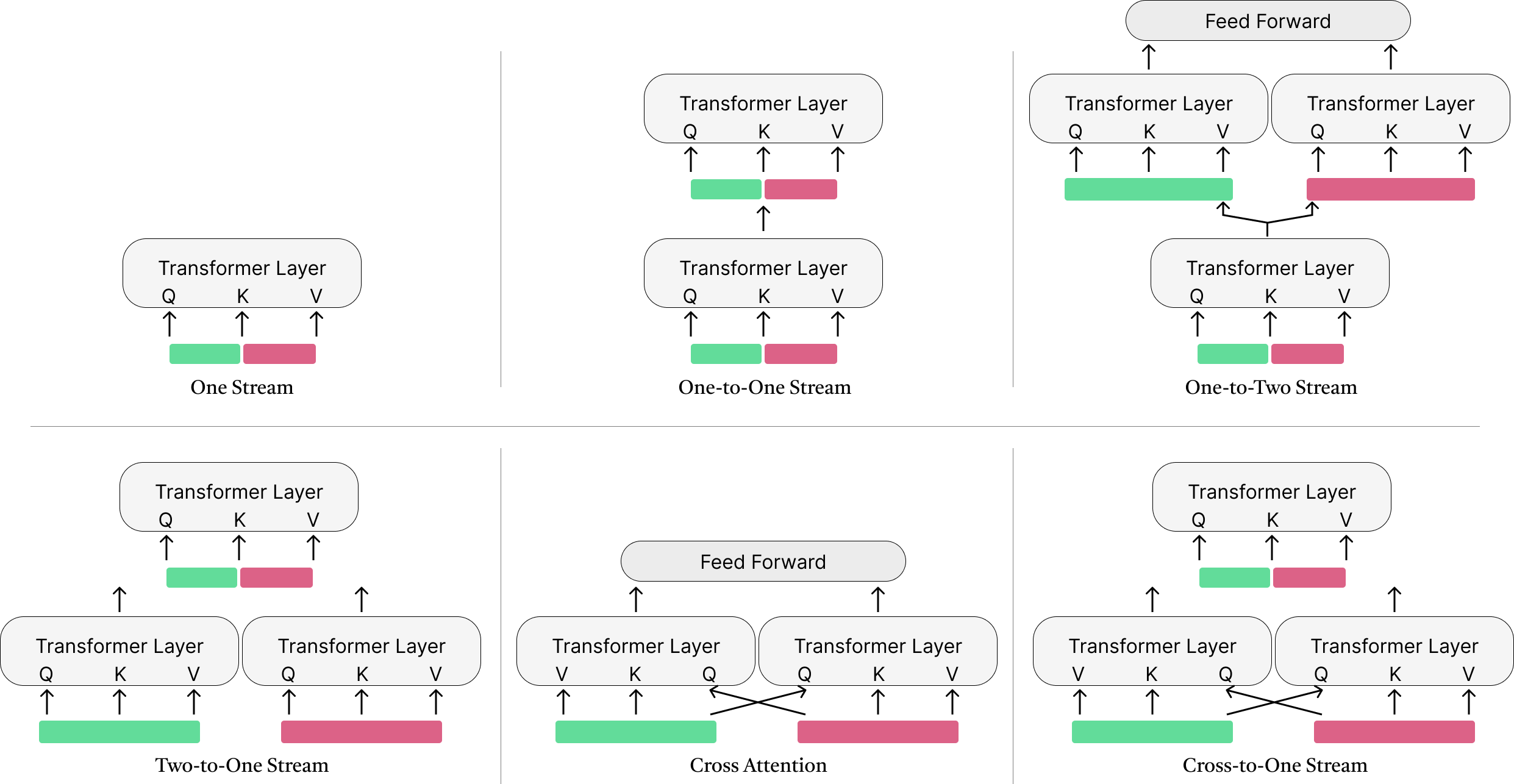}
  \caption{Transformer architectures for multi-modal fusion that were evaluated in this study. Different colors represent different modalities, which correspond to different sets of features in the study. {\textit{Q}}: Query embedding; {\textit{K}}: Key embedding; {\textit{V}}: Value embedding. Best viewed in colour.}
  \label{fig:multimodal_fusion_techniques}
\end{figure*}

\hfill

\subsection{Multi-Modal Fusion Transformer}

We evaluate all architectures presented in the recent survey on multi-modal transformers by Xu et al.~\cite{MultimodalTransformersSurvey} and Figure \ref{fig:multimodal_fusion_techniques} illustrates the different transformer architectures investigated in our work. However, we do not consider the early summation of feature vectors as this approach is not applicable to our scenario due to a mismatch in the feature dimensions.
In the following, we present each architecture from Figure \ref{fig:multimodal_fusion_techniques} in detail.

\textbf{\textit{One Stream}} is a multimodal fusion architecture that concatenates token embedding sequences from multiple modalities and inputs them into a transformer layer. This allows for the encoding of the global multimodal context, but increases computational complexity. The output of the transformer layer is then passed to a fully connected layer:
\begin{equation}
O = FC(TF(C(g_1(x),g_2(x)))) \label{eq:one-stream}
\end{equation}
where ${\textit{O}}$ is the output, ${\textit{FC}}$ is fully connected layer, ${\textit{TF}}$ is transformer layer and ${\textit{C}}$ stands for ${Concat()}$. $x$ is the input and $g_1$ and $g_2$ are two feature extracting transformations. 

\textbf{\textit{One-to-One stream}} is simply a vertical stacking of two \textbf{\textit{One Stream}} architectures. However, the fully connected layer is only used once after the second transformer layer:
\begin{equation}
O = FC({TF}_2({TF}_1(C(g_1(x),g_2(x))))) \label{eq:one-to-one-stream}
\end{equation}
This architecture is not discussed in the survey by Xu et al.~\cite{MultimodalTransformersSurvey}.
However, we include it as a comparison to the other two-layer architectures discussed in the survey article.

\textbf{\textit{One-to-Two stream}} is a hierarchical combination of a transformer layer that processes both modalities jointly (one-stream), with subsequent, modality-specific transformer layers (two-stream).
This allows for cross-modal interactions while maintaining the independence of uni-modal representations that can then be processed further by a subsequent linear layer. This splitting is intended to encode the different modalities separately which could possibly prevent confounding information from one modality corrupting the other in the intermediate feature space. The associated equations for this architecture are:
\begin{equation}
\begin{split}
{g_1}',{g_2}' = S({TF}_1(C(g_1(x),g_2(x)))) \\
O_1 = FC({TF}_2({g_1}')) \\
O_2 = FC({TF}_3({g_2}')) \\
O = FF(C(O_1,O_2)) \label{eq:one-to-two-stream}
\end{split}
\end{equation}
where ${\textit{S}}$ means splitting the output to 2 separate channels intended to separate the input modalities in the intermediate feature space, ${{g_1}'}$ and ${{g_2}'}$ stand for the separated channels in the intermediate space and ${\textit{FF}}$ denotes a feed forward neural network.

\textbf{\textit{Two-to-One stream}} reverses the layer ordering of the one-to-two stream architecture. 
The input modalities are encoded with independent Transformer layers, their outputs concatenated, and then fed through a one-stream Transformer layer. The motivation is to let the final transformer layer ascertain the interactions and confounding information supplied by the transformed intermediate input representations. The equation for this architecture is:
\begin{equation}
O = FC({TF}_3(C({TF}_1(g_1(x)) , {TF}_2(g_2(x))))) \label{eq:two-to-one-stream}
\end{equation}

\textbf{\textit{Cross Attention}} is a technique used in two-stream Transformers to perceive cross-modal interactions by exchanging ${\textit{Q}}$ (Query) embeddings. %
\cite{MultimodalTransformersSurvey} shows that two-stream cross-attention can effectively learn cross-modal interactions, but does not incorporate self-attention to the self-context within each modality. The equation for this architecture is:
\begin{equation}
O = FC(C({{TF}_{1}}'(g_1(x)) , {{TF}_{2}}'(g_2(x)))) \label{eq:cross-attention}
\end{equation}
where $TF_{1}'$ and $TF_{2}'$ are the transformer models with the interchanged $Q$ embeddings.

\textbf{\textit{Cross-to-one Stream}} architecture is concatenating the outputs of the two streams of cross-attention, then inputting it into an additional Transformer layer to capture the global context. The equation for this architecture is:

\begin{equation}
O = FC(TF_3(C({{TF}_{1}}'(g_1(x)) , {{TF}_{2}}'(g_2(x))))) \label{eq:cross-attention-two-to-one}
\end{equation}

\subsection{Features} \label{subsection:method-features}

In this study, we extracted facial and body keypoints using OpenFace 2.0 \cite{baltrusaitis2018openface} and OpenPose \cite{openpose} respectively, to be used as features from the available video information \cite{muller2022multimediate}. The transforms $g_1$ and $g_2$ used in Equations [\ref{eq:one-stream} - \ref{eq:cross-attention-two-to-one}] are representing the OpenPose Features $p=g_2(x) \in \mathbb{R}^{76}$ and OpenFace 2.0 Features $f=g_1(x) \in \mathbb{R}^{674}$ respectively.
Figure \ref{fig:openface-openpose} shows a sample from the group interaction dataset on the left, OpenFace 2.0 features applied on the sample in the middle, and OpenPose features applied on the sample on the right.  
The OpenFace 2.0 features include facial attributes, such as gaze direction, head pose, and facial action unit (AU) intensities, which denote the degree of activation of facial muscles.
For OpenPose features ${\textit{p}}$, only the skeleton pose key points (no facial key points) were utilized.
Preliminary experiments revealed an improvement in performance when including the frame number at each position of the input sequence.
In this study, we used the absolute difference of features between each consecutive frame as an input in order to directly encode the movement dynamics. 
This proved superior over inputting raw features or signed feature differences in preliminary experiments.

\section{Experimental Evaluation}

\subsection{Dataset}
For our evaluations we make use of the MPIIGroupInteraction dataset~\cite{muller_detecting_2018}.
This dataset comprises of 24 group discussions on controversial topics.
Groups consisted of three to four participants each and discussions lasted for 20 minutes.
All participants were recorded with frame-synchronized cameras.
The dataset was later annotated for backchannels as well as the amount of agreement expressed in each backchannel~\cite{muller2022multimediate}.
Both verbal- and nonverbal backchanneling behavior was annotated. 
For comparability, we precisely follow the definition of the backchannel detection and agreement estimation tasks used in the MultiMediate challenge, including the training and validation splits~\cite{muller2021multimediate,muller2022multimediate}.
In the MultiMediate challenge, each sample consists of a 10 second long video and a corresponding ground truth annotation.
In the backchannel detection task, 3358
training (including 1427 validation) samples contain a backchannel in the last second of the input video.
The same number of samples without a backchannel during the last second is provided as the negative class.
In the agreement estimation task on the other hand, only the samples containing backchannels during the last second of the input video are provided.
The ground truth in this regression task is the amount of agreement this backchannel expresses towards the current speaker.

\subsection{Implementation Details}

While the MultiMediate challenge provides 10 second input videos, only the last second may contain the relevant backchannel.
In preliminary experiments, we determined that utilizing the last 3 seconds of the input video resulted in the best performance.
Shorter intervals appear to not contain enough context and larger intervals tend to introduce a high amount of non task-related information.

The number of attention heads applied to the transformer that had OpenFace 2.0 features ${\textit{f}}$ as input was $4$, while the number of attention heads applied to the transformer that had OpenPose features ${\textit{p}}$ as input was $2$. The number of attention heads applied to the concatenated OpenFace 2.0 and OpenPose features as input was $10$. 
Positional encoding was applied only in the first layer of the transformer architectures.
This led to a slight improvement in performance compared to the application of positional encoding in all transformer layers.
For multi-layer transformer architectures, we furthermore utilized intermediate losses to directly supervise each transformer layer in addition to the supervision applied to the final output.
This led to a consistent improvement across different architectures.
For the intermediate loss equation, the weights were distributed as evenly as possible among the layers. The loss formula for One-to-Two Stream is shown in equation \ref{eq:loss-one-to-two-stream}.
\begin{equation}
\begin{split}
L_1 = 0.5*L_{TF_2} + 0.5*L_{TF_3} \\
L = 0.35*L_{TF_1} + 0.35*L_1 + 0.3*L_{FF}  \label{eq:loss-one-to-two-stream}
\end{split}
\end{equation}
Here, ${L_{TF_n}}$ is the loss of the $n^{th}$ transformer, with $n \in \{1,2,3\}$ for the 3 transformers used.

We utilized the Adam optimizer in all experiments, with a weight decay of $0.0005$ and a learning rate of $0.0005$. Binary Cross Entropy (BCE) loss was used for the backchannel detection task (classification), and the Mean Squared Error (MSE) loss for the agreement estimation task (regression).
We set the number of epochs to 350, and the best model was saved according to the best results on the validation set. 
All experiments were implemented in Pytorch \cite{pytorch}\footnote{Code is available at \url{https://git.opendfki.de/body_language/ijcnn23-backchannel-detection}}.

\subsection{Baseline and Metrics}

To evaluate the utility of transformer networks in comparison to recurrent neural networks, we used both unidirectional and bidirectional LSTMs as baselines.
To ensure a fair comparison, we configured the LSTM architecture to have a similar number of parameters as the One Stream architecture. A single layer of LSTM was utilized. We set the hidden size of the unidirectional LSTM to $1045$, while the bidirectional LSTM to $660$, to maintain a consistent relative number of parameters. 

We followed the evaluation metrics used in the MultiMediate challenge~\cite{muller2022multimediate}.
For backchannel detection, we used accuracy (ACC).
For agreement estimation from backchannels we used mean squared error (MSE).

\section{Results}

We first report results on the MultiMediate'22~\cite{muller2022multimediate} validation set to identify the best-performing transformer architecture and to compare against LSTM baselines.
Subsequently we report results on the MultiMediate'22 test set to evaluate our improvements over the state-of-the-art on backchannel detection and agreement estimation from backchannels.

\subsection{Comparison of Transformer Architectures}

We present the results of our evaluation of different transformer architectures in \autoref{table:ablation-results-validation}.
For the backchannel detection task, the best validation performance was achieved by a single-layer, one-stream transformer network (0.736 ACC).
The second best approach was a single cross-attention layer with (0.732 ACC).
All transformer-based approaches outperformed both the official challenge baseline (0.639 ACC) as well as uni- and bidirectional LSTM networks (0.575 and 0.692 ACC).
The results obtained when single modalities were used as inputs were inferior to the official challenge baseline. The accuracy achieved with OpenFace 2.0 features and OpenPose features was (0.520 and 0.597 ACC), respectively.

In the case of agreement estimation from backchannels, the best performance (0.0644 MSE) was achieved by two subsequent single-stream transformer layers, followed by the one-to-two stream variant (0.0668 MSE).
The single-layer, one-stream transformer which performed best for backchannel detection, reached an MSE of (0.0672) for agreement estimation.
All transformer-based approaches outperform uni- and bidirectional LSTM baselines (0.0871 and 0.0792 respectively).
Only the dual-to-one stream transformer model performs slightly worse than the official challenge baseline (0.0750 MSE).
In contrast to backchannel detection, the one-stream model relying solely on facial features only suffers from a minor reduction in performance compared to the one stream model integrating face- and pose features (0.0677 MSE vs. 0.0672 MSE).
In contrast, relying on pose features only led to a large reduction in performance (0.0840 MSE).

\begin{table}
\caption{\label{table:ablation-results-validation} Validation results for backchannel detection and agreement estimation from backchannels on different fusion techniques, figure \ref{fig:multimodal_fusion_techniques}.} 
\centering
\begin{tabular}{p{0.55\linewidth} | p{0.15\linewidth} | p{0.15\linewidth}}
\hline \textbf{Approach} & \textbf{Detection Acc ↑} & \textbf{Agreement MSE ↓} \\ \hline
One Stream & \textbf{0.736} & 0.0672 \\
One-to-One Stream & 0.728 & \textbf{0.0644} \\
One-to-Two Stream & 0.725 & 0.0668 \\
Two-to-One Stream & 0.709 & 0.0754 \\
Cross Attention & 0.732 & 0.0711 \\
Cross-to-One Stream & 0.711 & 0.0698 \\
\hline
One Stream (face only) & 0.520 & 0.0677 \\
One Stream (pose only) & 0.597 & 0.0840 \\
\hline
Unidirectional LSTM & 0.575 & 0.0871 \\
Bidirectional LSTM & 0.692 & 0.0792 \\
Sharma et al. \cite{sharma2022graph} & 0.693 & 0.073 \\
Challenge Baseline (SVM) \cite{muller2022multimediate} & 0.639 &  0.0750 \\
\hline
\end{tabular}
\end{table}

\subsection{State-of-the-Art Comparison}

For evaluation against the state of the art in backchannel detection and agreement estimation, we submitted the predictions of our models with the highest validation performance to the MultiMediate challenge organizers for evaluation on the test set.
\autoref{table:sota-results-backchanneldetection} shows the results for the backchannel detection task. Included in the table are also approaches that have not yet been published.
Our approach reached a test accuracy of 0.664, outperforming all published (0.621 ACC, \cite{sharma2022graph}) and unpublished competitors (0.658 ACC).
It also improved over the official challenge baseline (0.596 ACC) by a large margin.
Results for the agreement estimation task are shown in \autoref{table:sota-results-agreementestimation}.
As in the case of backchannel detection, our transformer-based method achieves the highest performance (0.0604 MSE) of all evaluated approaches.
At the same time, the margin of improvement over the official challenge baseline (0.0609) is rather small.
We would like to note however that our method is the first that was able to improve over the official challenge baseline at all.
For example, the approach of Sharma et al.~\cite{sharma2022graph} reached only 0.0623 MSE even though it led to improvements in the backchannel detection task.
This illustrates the extremely challenging nature of the agreement estimation task.

\subsection{Qualitative Results}

\begin{figure}
    \centering
    \includegraphics[width=\linewidth]{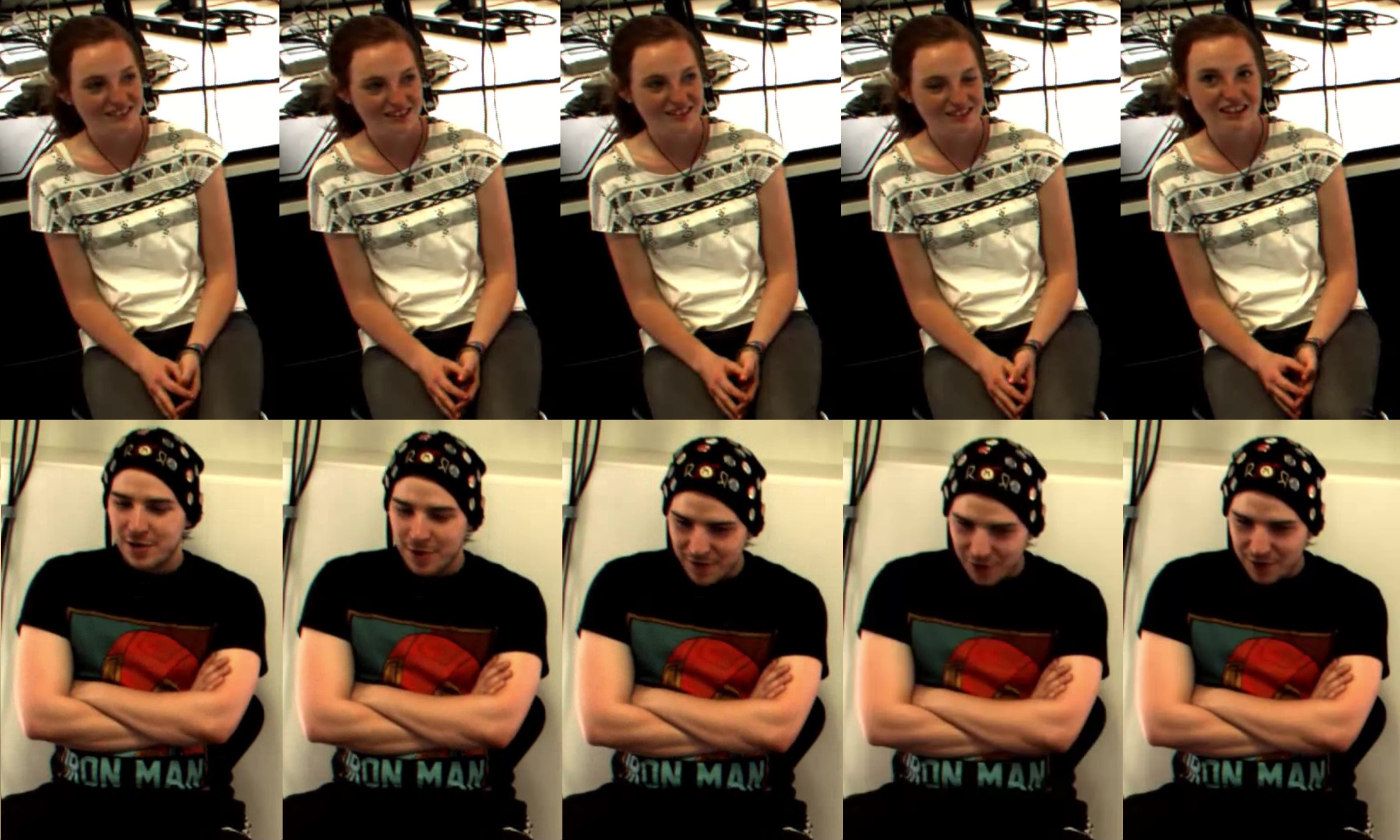}
    \caption{Qualitative examples for the backchannel detection task. \textbf{Top:} Ground truth backchannel correctly predicted by our method. \textbf{Bottom:} Ground truth backchannel missed by our method. The figures were magnified relative to the original to show the facial features of the participants.}
    \label{fig:best-worse-cases-backchannel-detection}
\end{figure}

We used the best model for each task to show qualitative results.
Figure \ref{fig:best-worse-cases-backchannel-detection} illustrates the results of the backchannel detection task. The top image represents a correctly predicted backchannel.
The participant is smiling and nodding, which are two common facial backchannel cues. 
The bottom image represents an incorrect prediction, where the model failed to detect the presence of a backchannel.
Here, the participant only slightly tilted their head down, which is a more subtle backchannel cue compared to the first example.

\begin{figure}
    \centering
    \subfigure{\includegraphics[width=.28\linewidth]{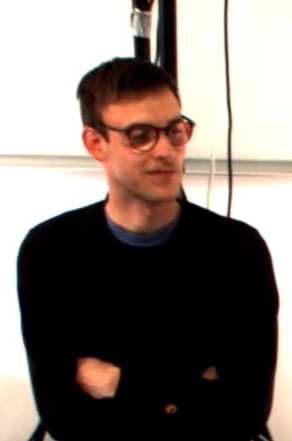}}
    \subfigure{\includegraphics[width=.3\linewidth]{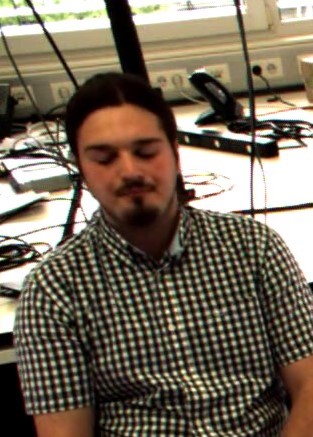}}
    \subfigure{\includegraphics[width=.28\linewidth]{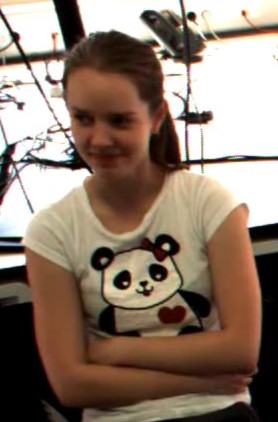}}
    \caption{Qualitative examples for the agreement estimation task. \textbf{Left:} perfect prediction of our model (Ground Truth: 0.25, Prediction: 0.25), \textbf{Middle:} wrong prediction (Ground Truth: 0.917, Prediction: 0.148), \textbf{Right:} (Ground Truth: -0.75, Prediction: 0.134). The figures were magnified relative to the original to show the facial features of the participants.}
    \label{fig:best-worse-cases-agreement-estimation}
\end{figure}

For the agreement estimation task, Figure \ref{fig:best-worse-cases-agreement-estimation} displays a perfectly predicted sample on the left, while the other two are incorrect predictions. In the left image, both the original and predicted agreement estimations are (0.25), indicating slight agreement from the participant. The participant is slightly smiling and nodding, which led to a correct prediction by the model. In the middle image, the original agreement estimation is (0.917), while the prediction is (0.145). 
While the participant is nodding, the facial expression shown in this example is more associated with a negative evaluation, which appears to have misled the model.
On the right, the original agreement estimation is (-0.75), indicating a high level of disagreement, while the prediction is (0.133), indicating agreement. This discrepancy is likely due to the participant smiling, which may have confused the model into predicting agreement.

\begin{table}[t]
\caption{\label{table:sota-results-backchanneldetection} Backchannel Detection Leaderboard of the MultiMediate Challenge\tablefootnote{\url{https://multimediate-challenge.org/leaderboards/leaderboard_backchannel/}}}
\centering
\begin{tabular}{p{0.65\linewidth} | p{0.2\linewidth}}
\hline \textbf{Username and Affiliation (Backchannel Detection)} & \textbf{Test Accuracy} \\ \hline
\textbf{Ours (One Stream)} & \textbf{0.664} \\
Anonymous 1 (publication pending) & 0.658 \\
Ma et al. (publication pending) & 0.656 \\
Sharma et al. \cite{sharma2022graph} & 0.621 \\
Baseline 2022: Head + Pose Features \cite{muller2022multimediate} & 0.596 \\
Baseline 2022: All Features \cite{muller2022multimediate} & 0.592 \\
Baseline 2022: Trivial (most likely class) \cite{muller2022multimediate} & 0.500 \\
\hline
\end{tabular}
\end{table}

\begin{table}[t]
\caption{\label{table:sota-results-agreementestimation} Backchannel Agreement Estimation Leaderboard of the MultiMediate Challenge\tablefootnote{\url{https://multimediate-challenge.org/leaderboards/leaderboard_backchannel-agreement/}}}
\centering
\begin{tabular}{p{0.65\linewidth} | p{0.2\linewidth}}
\hline \textbf{Username and Affiliation (Agreement Estimation)} & \textbf{Test MSE} \\ \hline
\textbf{Ours (One-to-One Stream)} & \textbf{0.0604} \\
Baseline 2022: Head Pose Features only \cite{muller2022multimediate} & 0.0609 \\
Sharma et al. \cite{sharma2022graph} & 0.0623 \\
Baseline 2022: All Features \cite{muller2022multimediate} & 0.0643 \\
Ma et al. (publication pending) & 0.0650 \\
Baseline 2022: Trivial (mean on train) \cite{muller2022multimediate} & 	0.0665 \\
\hline
\end{tabular}
\end{table}

\section{Discussion}

\subsection{Achieved Performance}

First and foremost, our results document the impressive effectiveness of transformer-based architectures.
In line with recent work in computer vision~\cite{CVTransformer} and natural language processing~\cite{AttentionIsAllYouNeed}, our transformer networks clearly outperformed classical approaches as well as commonly used LSTM networks.
They also set a new state of the art for backchannel detection and agreement estimation on the MultiMediate challenge~\cite{muller2022multimediate}.
In the case of backchannel detection, we improved with a clear margin over previously published approaches.
On the challenging task of agreement estimation from backchannels, our model improved slightly over the official challenge baselines, however it is the first approach that was able to do so at all.

Our experiments with single input feature modalities (face features only, pose features only) underline the crucial importance of using both face and pose information for backchannel detection.
In the agreement estimation task on the other hand, face features are dominating and there is only a slight improvement when adding pose features.
Interestingly, the best-performing architecture for backchannel detection consisted of a single transformer layer.
While more elaborate multi-modal fusion approaches like cross attention also improved over the previous state of the art on the validation set, they could not outperform the single-layer one-stream network.
This points to the importance of evaluating sophisticated fusion methods against simpler ones - something that is often not done in multi-modal transformer architectures~\cite{multimodalEmotionRecogTransformer,PersonalityAndBodyLanguageRecogTransformer, ValenceArousalEstimationMMTransformer,DyadFormer}.

\subsection{Limitations and Future Work}

While we established a new state-of-the-art for backchannel detection and agreement estimation from backchannels on the MPIIGroupInteraction corpus, it remains an open question to what extent our results would generalize to other scenarios.
At present, MPIIGroupInteraction is the only available corpus that allows to address both prediction tasks from video input.
As this corpus is limited to discussions in German, it will be important to collect further corpora, ideally involving participants speaking different languages and from different cultural backgrounds.

In our work we set out to address the backchannel detection and agreement estimation from video input only.
This has the advantage to be robust against bad quality- or even lack of audio recordings.
While in previous work the addition of the audio modality resulted in little or no improvement~\cite{muller2022multimediate,sharma2022graph}, it will be worthwhile for future work to investigate this issue in the scenario of multi-modal transformer architectures as well.

Finally, while the detection of- and the estimation of agreement from backchannels is a crucial step towards automatic conversation analysis, future work should investigate how to best make use of this information.
Automatic backchannel analysis could potentially be useful in systems that manage engagement~\cite{schiavo2014overt}, to better characterise states of psychiatric patients~\cite{konig2022multimodal}, or as a feature for exploratory conversation analysis~\cite{penzkofer2021conan}.

\section{Conclusion}

In this paper, we conducted a comprehensive evaluation of multi-modal transformer architectures for automatic backchannel analysis from body pose and facial behaviour.
We identified a single one-stream transformer layer as the best-performing architecture for backchannel detection, outperforming the previous state of the art on the MultiMediate challenge.
For agreement estimation, an architecture consisting of two subsequent one-stream transformer layers performed best, reaching a new state of the art on this task.
Our ablation and baseline comparison experiments, prove the general effectiveness of transformer-based approaches over recurrent neural networks for automatic backchannel analysis, but also point to the need to carefully evaluate sophisticated architectures against more simple ones that might reach competitive- or even better performance.

\section*{Acknowledgments}
This research was funded by the German Ministry for Education and Research (BMBF; grant number 01IS20075).

\bibliographystyle{IEEEtran}
\bibliography{bibliography}

\end{document}